\documentclass[runningheads]{llncs}
\usepackage[T1]{fontenc}
\usepackage[colorlinks,linkcolor=blue]{hyperref}
\usepackage{booktabs}
\usepackage{multirow}
\usepackage[T1]{fontenc}
\usepackage{amsfonts}
\usepackage{amsmath}
\usepackage{float}
\usepackage{makecell}
\usepackage{xcolor}
\usepackage{marvosym} 
\hypersetup{hidelinks} 
\usepackage{graphicx,verbatim}
\usepackage{cite}
\begin{document}
%
\title{DynFS-MoE: Dynamic Functional-Structural Mixture-of-Experts for Post-Traumatic Epilepsy Diagnosis}
 \titlerunning{Dynamic Functional-Structural Mixture-of-Experts Model for PTE}
%

\author{
Jun-En Ding\inst{1,2} 
\and
Spencer Chen\inst{3} 
\and
Henry Noren\inst{3} 
\and
Daniel Valdivia\inst{3} 
\and
Christine Yohn\inst{3} 
\and
Suhina Patel\inst{3} 
\and
Taylor Zink\inst{3} 
\and
Hai Sun\inst{3} 
\and
Feng Liu\inst{1,2}\textsuperscript{(\Letter)} 
}

\authorrunning{J.-E. Ding et al.}
\institute{Department of Industrial and Systems Engineering, Rutgers University, \\
Piscataway, NJ, USA\\
  \and  
  Department of Systems Engineering, Stevens Institute of Technology, \\ Hoboken, NJ, USA\\ 
  \and
  Department of Neurosurgery, Robert Wood Johnson Medical School, Rutgers University, New Brunswick, NJ, USA\\
  \email{\{junen.ding,feng.liu\}@rutgers.edu}
}
  
\maketitle              
\begin{abstract}

Post-traumatic epilepsy (PTE) is a severe complication of traumatic brain injury (TBI). Yet, early identification remains challenging due to the complex structural and functional alterations it induces in the brain. To address this, we propose a dynamic multimodal Mixture-of-Experts (MoE) framework that integrates functional and structural connectivity through time-aware functional-structural encoding and class-conditioned expert routing. Within this framework, modality-specific and cross-modal experts learn complementary representations, while a Modality-Class MoE (MCoE) module dynamically adjusts expert weights according to each classification objective. Experimental results across three binary classification tasks demonstrate that the framework consistently outperforms static fusion baselines, and interpretability analyses further reveal meaningful interactions among regions of interest (ROIs). This dynamic multimodal expert framework effectively captures class-dependent brain interaction patterns and provides an interpretable approach for PTE diagnosis and risk stratification.


\end{abstract}

\section{Introduction}

Traumatic brain injury (TBI) remains a leading cause of morbidity and mortality worldwide, affecting millions annually and imposing substantial socioeconomic burdens~\cite{dewan2018estimating,injury2019global}. Among its long-term complications, TBI carries a significant latent risk of progression to post-traumatic epilepsy (PTE), which is associated with high recurrence rates and, in severe cases, may lead to mortality. PTE is characterized by recurrent unprovoked seizures occurring more than seven days after injury~\cite{acosta2024update,yang2025post,yang2026large}.

Currently, structural and functional connectivity offer high spatial resolution and network-level insights that effectively complement electroencephalogram (EEG) limitations. Structural MRI can precisely detect microstructural damage, brain atrophy, and hippocampal or thalamic volume loss~\cite{mohamed2021traumatic,akrami2022neuroanatomic,immonen2019imaging}. Moreover, resting-state functional magnetic resonance imaging (fMRI) further reveals functional connectivity abnormalities associated with the transition from TBI to PTE, such as thalamocortical hyperconnectivity, excessive integration, and reduced segregation patterns, thereby capturing the network mechanisms underlying epileptogenesis~\cite{la2023functional,garner2019imaging}.

However, most existing studies employ only structural MRI~\cite{akrami2022neuroanatomic,immonen2019imaging} or fMRI individually~\cite{akrami2024prediction}, focusing on structural damage (such as lesion volume and hippocampal or thalamic atrophy) or functional connectivity abnormalities (such as hyperconnectivity patterns)~\cite{la2023functional,garner2019machine}. Although these studies reveal significant differences in specific brain regions (e.g., temporal lobe and cerebellum), unimodal models generally yield lower predictive accuracy compared to multimodal integration and cannot simultaneously capture temporal dynamics and spatial network information, resulting in insufficient explanatory power for the heterogeneous progression of PTE~\cite{fang2023advances,storti2012multimodal,ebrahimzadeh2022simultaneous,akbar2024advancing}. 

Emerging evidence suggests that multimodal integration can substantially improve PTE prediction accuracy. For instance, machine learning models combining structural MRI, functional MRI, and EEG have achieved AUC values above 0.78~\cite{akrami2024prediction}. However, multimodal fusion faces several fundamental challenges. First, most fusion strategies project all modalities into a single shared feature space, potentially leading to information loss. Second, modality competition may arise during fusion, where dominant modalities overshadow weaker ones, and inter-individual variability in modality informativeness is often neglected. Finally, different TBI or PTE phenotypes may benefit from distinct primary modalities, necessitating adaptive rather than uniform weighting of modality contributions. To address the above limitations, we propose three modules to handle the challenges of dynamic fMRI and diffusion MRI (dMRI)-derived structural connectivity integration. Our contributions are as follows: (1) We design a temporal MoE mechanism that leverages structural connectivity to guide the selection of the most informative 
patches within fMRI temporal segments for cross-modal fusion and routing. (2) Modality-specific experts (MSoEs) are employed to extract individual feature representations from each modality, producing dedicated embeddings that preserve modality-unique information. (3) A class-based MoE module adaptively selects the most class-relevant modality patches to facilitate more discriminative classification.

\begin{figure}
    \centering
    \includegraphics[width=0.8\textwidth]{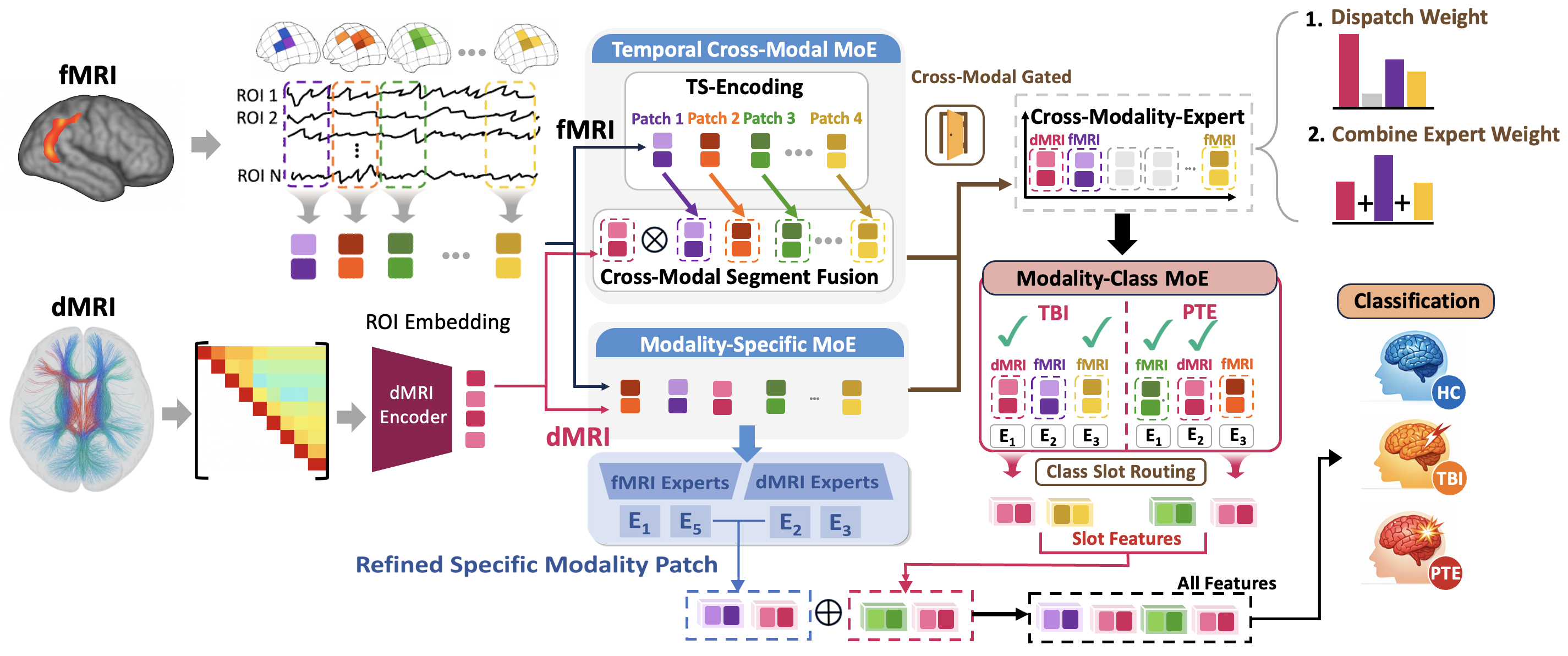}
    \caption{Overview of the proposed DynFS-MoE framework.}
    \label{fig:main_flow}
\end{figure}

\section{Methodology}

As illustrated in Fig.~\ref{fig:main_flow}, \textbf{DynFS-MoE} consists of three components: (1) a time-aware functional-structural encoder that extracts temporal fMRI and structural connectivity representations into refined modality patches through segment-wise fusion; (2) a mixture-of-experts layer composed of modality-specific experts and cross-modal experts, which capture both modality-specific characteristics and cross-modal interactions; and (3) a modality-class MoE (MCoE) routing module that dynamically assigns expert weights using a gating mechanism conditioned on class-aware representations. The routed expert outputs are combined to select the most suitable multimodal features for classification.

\subsection{Expert Space and Routing Formulation}

We define the expert set as $\mathcal{E} = \{e_1, e_2, \dots, e_J\}$, where $J$ denotes the total number of experts and each expert is parameterized as a learnable transformation $\phi_j : \mathbb{R}^{D} \rightarrow \mathbb{R}^{D}$. The expert set comprises modality-specific experts, which specialize in processing representations from either fMRI or dMRI, and cross-modal experts, which capture interactions between functional and structural modalities. To formalize the routing process, we introduce three discrete random variables: the modality patch variable $P \in \mathcal{P}$, the expert variable $E \in \mathcal{E}$, and the diagnostic label $Y \in \{1, \dots, C\}$, where $C$ denotes the number of diagnostic categories. The routing mechanism is modeled as a class-conditioned joint distribution $P_\theta(P, E \mid Y)$, parameterized by $\theta$, which defines the probability of assigning modality patches to experts conditioned on the class-specific representation. This probabilistic formulation enables class-aware expert selection and adaptive integration of modality-specific and cross-modal information, allowing the model to dynamically allocate expert capacity according to diagnostic relevance.

\subsection{Time-Aware Functional-Structural MoE Encoder}

Throughout this paper, we denote fMRI BOLD signals  $\textbf{F} \in \mathbb{R}^{R \times T}$ and the structural connectivity matrix, denoted as $\textbf{S} \in \mathbb{R}^{R \times R}$, where $R$ is the number of regions of interest (ROIs) and $T$ is the number of time steps. To jointly model the temporal dynamics of fMRI and the structural information from dMRI, we propose a time-aware functional-structural encoder that converts raw multimodal inputs into refined modality-specific patches. Our goal is to learn  a classifier $f_{\theta}: (\textbf{F},\textbf{S}) \to \mathcal{Y}$, parameterized by $\theta$, which predicts the diagnostic label $y \in \mathcal{Y}= \{1,...,C\}$. 

\subsubsection{Modality Patch-Wise Representation.}

To explicitly model temporal dynamics, the fMRI sequence is partitioned into $K = \left\lfloor \frac{T}{L} \right\rfloor$ non-overlapping temporal segments of fixed length $L$ as patches, where each modality patch represents an embedding of modality-specific temporal (fMRI) or structural connectivity information. Specifically, the fMRI temporal patches $\textbf{F}=\{F^{(1)},F^{(2)},...,F^{(K)}\}$ with each $F^{(k)}$ $\in \mathbb{R}^{R \times L}$. For the structural modality, we obtain a compact representation by extracting the upper-triangular entries of the connectivity matrix, denoted as $\mathrm{vec}_\triangle(S) \in \mathbb{R}^{\frac{R(R+1)}{2}}$, where $\mathrm{vec}_\triangle(\cdot)$ collects elements $S_{i,j}$ satisfying $1 \leq i \leq j \leq R$. We then encode temporal patches and structural patches to MLP layer $\phi(\cdot)$ as same dimensional embedding $f^{(k)} \in \mathbb{R}^{D}$ and  $s \in \mathbb{R}^{D}$, respectively. The projected unified modality patch space is $\mathcal{P} = \{f^{(1)}, \dots, f^{(K)}, s\}$, with cardinality $|\mathcal{P}| = K + 1$. This unified patch formulation enables both temporal functional patches and structural connectivity features to be processed within the same MoE routing framework described in Section~\ref{sec:MSoE} and~\ref{sec:MCMoE}, facilitating structured expert assignment and cross-modal integration.

\subsubsection{Cross-Modal Interaction Fusion.}\label{sec:cross-modal}
Unlike previous approaches that adopt a traditional unified modality
paradigm, we propose a lightweight cross-modal fusion to integrate
structural connectivity features into each fMRI temporal patch prior to
MoE routing. For each temporal patch, a scaled dot-product gating
is performed as:

\begin{equation}
\tilde{f}^{(k)} = f^{(k)} + \alpha^{(k)} \mathbf{W}_V s,
\end{equation}
where the gating coefficient $\alpha^{(k)} =
\sigma\!\left(
\frac{
\left(\mathbf{W}_Q f^{(k)}\right)^\top
\left(\mathbf{W}_K s\right)
}{\sqrt{D}}
\right) \in (0,1)$ is a scalar sigmoid gate that adaptively
modulates the contribution of structural connectivity to each
functional patch. Next, a feed-forward projection is applied to
produce the cross-modal fused embedding $z_{fused}^{(k)}$ with
MLP $\phi(\tilde{f}^{(k)})$.

\subsection{Modality-Specific MoEs (MSoEs)}\label{sec:MSoE}

To capture modality-specific characteristics, we employ modality-specific MoEs operating on modality patch embeddings
$p_i \in \mathcal{P}, \; p_i \in \mathbb{R}^{D}$. Given the expert set $\mathcal{E}$, and
a gating function computes routing probabilities
\begin{equation}
\pi_j(p_i)
= P_\theta(E = e_j \mid P = p_i)
= \mathrm{softmax}_j\!\left(\mathbf{W}_g p_i\right).
\end{equation}
where $\pi_j(p_i)$ represents the routing probability of assigning patch $p_i$ to expert $e_j$, and $\sum_{j=1}^{J} \pi_j(p_i) = 1.$ We can calculate the modality-specific representation with a weighted combination of expert outputs $\mathbf{z}^{\mathrm{spec}}_i
=
\sum_{j=1}^{J}
\pi_j(p_i)\,\phi_j(p_i)$, which allows different experts to capture heterogeneous modality-specific patterns in functional dynamics and structural connectivity, thereby improving representation expressiveness prior to class-conditioned routing in the next Section~\ref{sec:MCMoE}

\subsection{Modality-Class MoE (MCoE) with Class-Conditioned Routing }\label{sec:MCMoE}

Inspired by existing work~\cite{wu2025dynamic}, we introduce the Modality-Class Mixture of Experts (MCoE) module, which enables different classes to learn the most suitable modality representations according to their assigned modality combinations and expert specialization. Specifically, the joint distribution $P_{\theta}(P,E|Y)$ allows different diagnostic classes, e.g., healthy control (HC), TBI, and PTE, to dynamically route modality patches to specialized experts, capturing heterogeneous modality dependencies.

Formally, the input to this module is the set of refined modality patch embeddings $  \mathbf{Z} = \{ \mathbf{z}_i \}_{i=1}^{|P|}  $, where each $  \mathbf{z}_i \in \mathbb{R}^D  $ incorporates cross-modal fusion and modality-specific processing from prior layers (Sections~\ref{sec:cross-modal} and ~\ref{sec:MSoE}). We compute class-conditioned routing logits as:

\begin{equation}
\mathbf{L} = f(\mathbf{Z}) \odot \Phi \in \mathbb{R}^{B \times |P| \times J \times C}.
\end{equation}
where $B$ denotes the batch size, and $  \Phi \in \mathbb{R}^{|P| \times J \times C} $ is the parameter matrix. Next, we design the dispatch weights $D_{i j y} = \frac{\exp(L_{i j y})}{\sum_{i'=1}^{|P|} \exp(L_{i' j y})}$, which aggregate patches for each expert per class, representing $  P(P_i \mid E_j, Y=y)  $. For each class $  y  $ and expert $  j  $, the expert input is a weighted sum augmented by class embedding $  \mathbf{c}_y \in \mathbb{R}^D  $ with the class-conditioned expert input $\mathbf{u}_{j y}$ and expert output $\mathbf{o}_{j y}$ can be expressed as:
\begin{equation}
    \mathbf{u}_{j y} = \left( \sum_{i=1}^{|P|} D_{i j y} \cdot \mathbf{z}_i \right) + \mathbf{c}_y, \quad \mathbf{o}_{j y} = \phi_j(\mathbf{u}_{j y}),
\end{equation}
To combine expert outputs into class-specific features, we compute combine weights  $C_{j y} = \frac{\exp(\bar{L}_{j y})}{\sum_{j'=1}^J \exp(\bar{L}_{j' y})}$ via softmax over experts, using logits from mean-pooled $  \mathbf{Z}  $, and we can  yield the fused feature $\textbf{h}$ for class $y$ as $ \mathbf{h}_y = \sum_{j=1}^J C_{j y} \cdot \mathbf{o}_{j y}$, where class-specific features $  \{\mathbf{h}_y\}_{y=1}^C  $ are concatenated with modality-specific representations and passed to a final MLP classifier for predicting $  y  $.

\subsection{Brain Interpretability Regularizer}

Our training aims to promote diagnostic relevance using class-conditioned routing, allowing expert capacity to be allocated adaptively. To encourage class-distinctive patterns, we add a brain interpretability regularizer maximizing conditional mutual information $  I(P; E \mid Y) = H(P \mid Y) - H(P \mid E, Y)  $. Using logits $  \mathbf{L}  $, joint and marginal probabilities are $P(P_i, E_j \mid Y) = \frac{\exp(L_{i j y})}{\sum_{i'=1}^{|P|} \sum_{j'=1}^J \exp(L_{i' j' y})},$ and class-conditioned mutual information (CMI) can be written as:

\begin{equation}
\begin{aligned}
&I_\theta(P; E \mid Y) \\
&= \sum_{y=1}^C P(Y=y) \sum_{i=1}^{|P|} \sum_{j=1}^J P_\theta(P_i, E_j \mid Y=y) \cdot \log \left( \frac{P_\theta(P_i, E_j \mid Y=y)}{P_\theta(P_i \mid Y=y) \cdot P_\theta(E_j \mid Y=y)} \right)
\end{aligned}
\end{equation}

Formally, the overall objective is to minimize the following loss over all model parameters $\theta$:

\begin{equation}
    \min_{\theta} \;\mathbb{E}_{(\mathbf{F},\mathbf{S},y) \sim \mathcal{D}} \Bigg[ \mathcal{L}_{\text{cls}}\big(f_\theta(\mathbf{F},\mathbf{S}), y\big) - \lambda \, I_\theta(P; E \mid Y) \Bigg]
\end{equation}
where $\mathcal{L}_{cls}$ is the cross-entropy loss, and $I_\theta(P; E \mid Y)$ is the CMI encouraging class-specific modality routing, and $\lambda$ is the balancing hyperparameter.

\vspace{-2mm}
\section{Experimental Setting and Results}

In this study, we acquired MRI data from participants, including HC (n=26), TBI (n=24), and PTE (n=19) using a 3T Siemens Prisma scanner, including diffusion-weighted imaging. Preprocessing was performed using SPM, MRtrix3, and FSL. T1 images were segmented and normalized to MNI space, and functional images were motion-corrected, co-registered, and normalized. Particularly, structural connectivity matrices were constructed from diffusion-weighted imaging via DTI tractography using the AAL atlas (116 regions)~\cite{tzourio2002automated}.

\begin{table}[h]
\centering
\color{black}
\caption{\textcolor{black}{Performance comparison across different classification tasks.
Statistical significance was evaluated between DynFS-MoE and the best remaining baseline using Welch's t-test
(* $p<0.05$, ** $p<0.01$, *** $p<0.001$).}}
\label{tab:group_comparison}
\scriptsize 
\begin{tabular}{|l|l|c|c|c|}
\hline
\textbf{Cohort} & \textbf{Model} & \textbf{ACC} & \textbf{AUC} & \textbf{F1} \\
\hline
\multirow{10}{*}{HC vs. TBI}
& fMRI-FC+MLP
& 0.76$\pm$0.19
& 0.72$\pm$0.28
& 0.62$\pm$0.35 \\
& dMRI+MLP
& 0.76$\pm$0.10
& 0.70$\pm$0.12
& 0.73$\pm$0.12 \\
& Fusion+GCN~\cite{kipf2016semi}
& 0.52$\pm$0.07
& 0.56$\pm$0.15
& 0.40$\pm$0.23 \\
& FBNetGen~\cite{kan2022fbnetgen}
& 0.54$\pm$0.08
& 0.56$\pm$0.05
& 0.25$\pm$0.30 \\
& EarlyFusion
& 0.76$\pm$0.10
& 0.74$\pm$0.12
& 0.70$\pm$0.12 \\
& IBGNN~\cite{cui2022interpretable}
& 0.68$\pm$0.12
& 0.67$\pm$0.18
& 0.65$\pm$0.16 \\
& IBrainGNN~\cite{QU2025103570}
& 0.60$\pm$0.09
& 0.46$\pm$0.23
& 0.41$\pm$0.25 \\
& AGIIBM~\cite{dong2025multi}
& 0.70$\pm$0.11
& 0.55$\pm$0.14
& 0.64$\pm$0.19 \\
& MultiViT~\cite{bi2023multivit}
& 0.74$\pm$0.10
& 0.71$\pm$0.14
& \textbf{0.76$\pm$0.07} \\
& Contrasformer~\cite{xu2024contrasformer}
& 0.72$\pm$0.07
& 0.60$\pm$0.09
& 0.73$\pm$0.06 \\
& \textbf{DynFS-MoE}
& \textbf{0.76$\pm$0.05}
& \textbf{0.77$\pm$0.09}
& 0.72$\pm$0.16 \\
\hline
\multirow{10}{*}{HC vs. PTE}
& fMRI-FC+MLP
& 0.78$\pm$0.10
& 0.73$\pm$0.13
& 0.73$\pm$0.11 \\
& dMRI+MLP
& 0.76$\pm$0.08
& 0.78$\pm$0.06
& 0.63$\pm$0.18 \\
& Fusion+GCN~\cite{kipf2016semi}
& 0.58$\pm$0.04
& 0.42$\pm$0.19
& 0.00$\pm$0.00 \\
& FBNetGen~\cite{kan2022fbnetgen}
& 0.62$\pm$0.09
& 0.44$\pm$0.08
& 0.25$\pm$0.31 \\
& EarlyFusion
& 0.71$\pm$0.09
& 0.65$\pm$0.18
& 0.51$\pm$0.16 \\
& IBGNN~\cite{cui2022interpretable}
& 0.60$\pm$0.05
& 0.70$\pm$0.27
& 0.18$\pm$0.22 \\
& IBrainGNN~\cite{QU2025103570}
& 0.62$\pm$0.09
& 0.39$\pm$0.26
& 0.16$\pm$0.32 \\
& AGIIBM~\cite{dong2025multi}
& 0.71$\pm$0.13
& 0.61$\pm$0.31
& 0.40$\pm$0.35 \\
& MultiViT~\cite{bi2023multivit}
& 0.78$\pm$0.07
& 0.70$\pm$0.10
& 0.74$\pm$0.06 \\
& Contrasformer~\cite{xu2024contrasformer}
& 0.80$\pm$0.11
& 0.68$\pm$0.18
& 0.72$\pm$0.20 \\
& \textbf{DynFS-MoE}
& \textbf{0.84$\pm$0.05$^{*}$}
& \textbf{0.84$\pm$0.10$^{***}$}
& \textbf{0.81$\pm$0.08$^{***}$} \\
\hline
\multirow{10}{*}{TBI vs. PTE}
& fMRI-FC+MLP
& 0.72$\pm$0.06
& 0.67$\pm$0.13
& 0.69$\pm$0.10 \\
& dMRI+MLP
& 0.68$\pm$0.08
& 0.55$\pm$0.08
& 0.58$\pm$0.12 \\
& Fusion+GCN~\cite{kipf2016semi}
& 0.56$\pm$0.04
& 0.51$\pm$0.26
& 0.00$\pm$0.00 \\
& FBNetGen~\cite{kan2022fbnetgen}
& 0.58$\pm$0.03
& 0.41$\pm$0.21
& 0.08$\pm$0.16 \\
& EarlyFusion
& 0.67$\pm$0.15
& 0.65$\pm$0.12
& 0.60$\pm$0.19 \\
& IBGNN~\cite{cui2022interpretable}
& 0.61$\pm$0.04
& 0.57$\pm$0.10
& 0.42$\pm$0.35 \\
& IBrainGNN~\cite{QU2025103570}
& 0.56$\pm$0.04
& 0.55$\pm$0.24
& 0.00$\pm$0.00 \\
& AGIIBM~\cite{dong2025multi}
& 0.63$\pm$0.11
& 0.52$\pm$0.14
& 0.25$\pm$0.32 \\
& MultiViT~\cite{bi2023multivit}
& 0.70$\pm$0.11
& 0.59$\pm$0.13
& 0.50$\pm$0.20 \\
& Contrasformer~\cite{xu2024contrasformer}
& 0.59$\pm$0.10
& 0.31$\pm$0.18
& 0.37$\pm$0.21 \\
& \textbf{DynFS-MoE}
& \textbf{0.77$\pm$0.01$^{***}$}
& \textbf{0.82$\pm$0.07$^{***}$}
& \textbf{0.70$\pm$0.04} \\
\hline
\end{tabular}
\label{tab:table1}
\end{table}

\subsection{Classification Performance Results}

In this section, we evaluated \textbf{DynFS-MoE} on three binary classification tasks. 
As shown in Table~\ref{tab:table1}, \textbf{DynFS-MoE} achieves the best AUC across all three tasks and competitive ACC/F1, with the strongest improvements observed for HC vs. PTE and TBI vs. PTE. For HC vs. TBI, 
\textbf{DynFS-MoE} achieves an AUC of 0.77, surpassing the strongest baseline 
EarlyFusion (AUC=0.74) by a substantial margin. The performance gains are most 
pronounced in the clinically challenging HC vs. PTE and TBI vs. PTE tasks, where 
\textbf{DynFS-MoE} attains AUCs of 0.84 and 0.82, respectively. Notably, several 
graph-based methods (Fusion+GCN, FBNetGen) yield near-zero F1 scores on 
PTE-related tasks, suggesting that static fusion strategies fail to capture the 
heterogeneous multimodal patterns underlying epileptogenesis. For HC vs. PTE, 
\textbf{DynFS-MoE} further achieves statistically significant improvements in both 
AUC and F1 ($p < 0.001$), with F1=0.81 outperforming the best baseline MultiViT 
(F1=0.74). For TBI vs. PTE, \textbf{DynFS-MoE} also yields significant gains 
($p < 0.001$) with low variance (ACC=$0.77{\pm}0.01$, AUC=$0.82{\pm}0.07$), 
demonstrating stable and reliable performance under dynamic multimodal routing.

\subsection{Ablation Studies}

\subsubsection{Class-Conditioned Routing Mechanism.}

We compare class-conditioned routing against a uniform prior baseline in an ablation study. As shown in Table~\ref{tab:ablation}, class-conditioned routing consistently improves AUC across all tasks, with pronounced gains in TBI vs. PTE (0.79 vs. 0.58) and HC vs. PTE (0.81 vs. 0.72), attributed to its ability to adaptively allocate expert capacity per diagnostic class, where PTE down-weights dMRI in favor of early functional patches while TBI up-weights mid-temporal fMRI and structural features. Although F1 variability may reflect class imbalance, the consistent AUC gains confirm that class-conditioned routing is essential for learning discriminative, disease-specific multimodal representations.

\vspace{-4mm}
\begin{table}
\centering
\caption{Ablation study on conditioning strategy}
\label{tab:ablation_E}
\scriptsize 
\begin{tabular}{llcc}
\hline
Cohort & Setting & AUC & F1 \\
\hline
\multirow{2}{*}{HC vs TBI}
& Uniform Prior 
& 0.78 $\pm$ 0.05 
& \textbf{0.73 $\pm$ 0.09} \\
& Class-Conditioned (proposed) 
& \textbf{0.79 $\pm$ 0.13} 
& 0.67 $\pm$ 0.20 \\
\hline
\multirow{2}{*}{HC vs PTE}
& Uniform Prior 
& 0.72 $\pm$ 0.16 
& \textbf{0.72 $\pm$ 0.13} \\
& Class-Conditioned (proposed) 
& \textbf{0.81 $\pm$ 0.04} 
& 0.58 $\pm$ 0.15 \\
\hline
\multirow{2}{*}{TBI vs PTE}
& Uniform Prior 
& 0.58 $\pm$ 0.13 
& 0.54 $\pm$ 0.29 \\
& Class-Conditioned (proposed) 
& \textbf{0.79 $\pm$ 0.12} 
& \textbf{0.71 $\pm$ 0.06} \\
\hline
\end{tabular}

\vspace{-15mm}
\label{tab:ablation}
\end{table}

\begin{figure}[t]
    \centering
\vspace{-10mm}
    \includegraphics[width=1\textwidth]{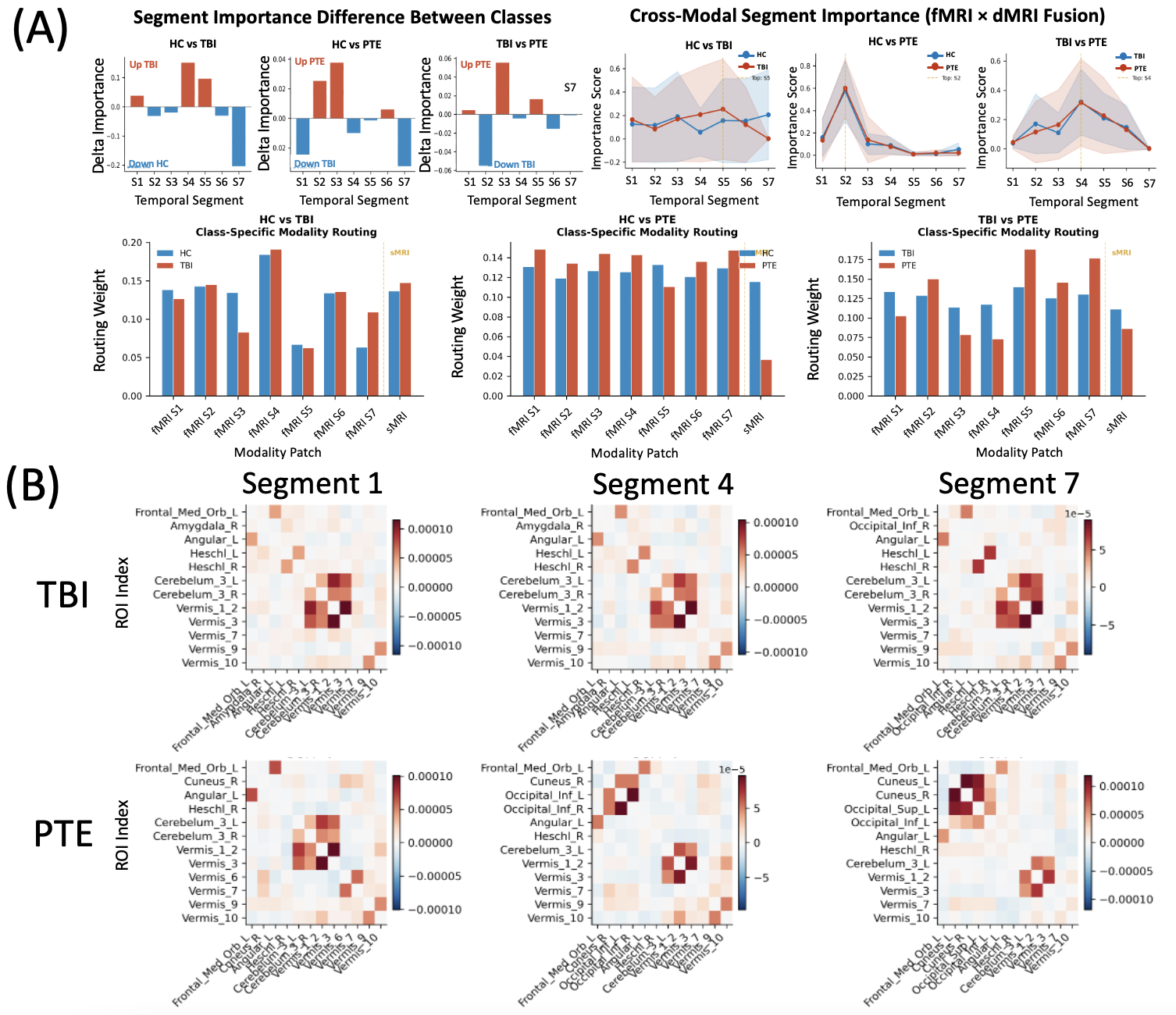}
    \caption{\textbf{Class-conditioned multimodal importance and ROI interaction visualization.}}
\vspace{-1mm}
    \label{fig:Fig2}
\end{figure}

\section{Functional Brain Interpretability and Structural Analysis}
Fig.~\ref{fig:Fig2} (A) and Fig.~\ref{fig:Fig2} (B) illustrate the learned cross-modal temporal importance in brain regions and structural connectivity patterns. Fig.~\ref{fig:Fig2} (A) depicts  three complementary interpretability analyses across the three binary classification tasks. PTE consistently exhibits elevated segment importance at early-to-mid temporal segments (notably S2--S4), with cross-modal importance curves peaking sharply at S2 in HC vs. PTE and S4 in TBI vs. PTE, suggesting that early functional-structural interactions are particularly informative for PTE identification. The class-specific modality routing further reveals that TBI up-weights mid-temporal fMRI patches and dMRI, HC distributes weights more uniformly, and PTE markedly down-weights dMRI in favor of functional patches, demonstrating adaptive modality reliance across diagnostic classes. Fig.~\ref{fig:Fig2} (B) displays pairwise ROI interaction heatmaps for TBI and PTE across Segments 1, 4, and 7. While both classes share a prominent cerebellar-vermis interaction core, PTE uniquely exhibits progressive recruitment of frontal-occipital (Frontal\_Med\_Orb\_L, Cuneus, Occipital\_Sup/Inf) and auditory cortical regions (Heschl\_R) in later segments, 
reflecting evolving functional-structural disruptions involving frontal-occipital, auditory, and cerebellar-vermis regions

\section{Conclusion}
In this study, we proposed DynFS-MoE, a dynamic multimodal mixture-of-experts framework for post-traumatic epilepsy (PTE) diagnosis after traumatic brain injury (TBI). By combining time-aware functional-structural encoding with class-conditioned expert routing, the model adaptively integrates fMRI and dMRI information and captures class-specific brain patterns. This dynamic routing strategy overcomes limitations of static fusion methods and improves discrimination performance across multiple classification tasks. Moreover, interpretability analyses reveal meaningful ROI interaction patterns, such as thalamic hyperconnectivity and limbic disruptions, providing clinically relevant insights. Overall, DynFS-MoE may provide an interpretable tool for PTE classification and could support future studies on early detection and risk stratification.

\section*{Funding Statement.}

This work was supported by the U.S. Dept of Defense and Congressionally Directed Medical Research Programs (CDMRP) under Award No. W81XWH-18-1-0655 (PI: Hai Sun).

\section*{Disclosure of Interest.}

All authors declare no potential conflict of interests.




%
%
%
\bibliographystyle{unsrt}
\bibliography{Ref}

\end{document}